\documentclass[runningheads]{llncs}
\usepackage[T1]{fontenc}
\usepackage{amsmath}
\usepackage{algorithm}
\usepackage{algorithmic}
\usepackage{booktabs}
\usepackage{multirow}
\usepackage{amssymb}
\usepackage{marvosym}
\usepackage{hyperref}

\usepackage{graphicx}
\usepackage{xcolor}

\begin{document}
\title{Eureka: Intelligent Feature Engineering for Enterprise AI Cloud Resource Demand Prediction}
\titlerunning{Eureka: Feature Engineering for Cloud Demand Prediction}
%

\author{Hangxuan Li\inst{12}\thanks{Hangxuan Li and Renjun Jia contributed equally to this work. This work was done during two author’s internship at Alibaba Cloud Computing Co. Ltd.} \and
Renjun Jia\inst{13}\protect\footnotemark[1] \and
Xuezhang Wu\inst{1}\and
Yunjie Qian\inst{1}\and
Zeqi Zheng\inst{1}\and
Xianling Zhang\inst{4}\Letter}

\authorrunning{H. Li et al.}
%
\institute{
Alibaba Cloud Computing Co. Ltd, Hangzhou, China\and
School of Computer Science, Fudan University, Shanghai, China\and
School of Computer Science and Technology, Tongji University, Shanghai, China\and
Independent Researcher, United States\\
\email{hxli23@m.fudan.edu.cn},
\email{2332101@tongji.edu.cn},
\email{\{weixi.wxz, ramsey.qyj, zhengzeqi.zzq\}@alibaba-inc.com},
\email{lilyzhng.ai@gmail.com}
}
\maketitle              
\begingroup 
\renewcommand\thefootnote{}
\footnotetext{\Letter\ Xianling Zhang is the Corresponding author.}
\endgroup

\begin{abstract}
\vspace{-0.1in}

Effective features are crucial for predictive model performance, but creating them often requires domain expertise, limiting scalability across applications. We define feature engineering as an \textbf{agentic code generation} problem: features are not static data transformations, but executable programs that can be generated, evaluated, and iteratively improved. We present \textbf{Eureka}, an LLM-driven framework with three stages. (1) An \textit{Expert Agent}, fine-tuned via SFT on domain knowledge, produces structured feature design plans in JSON format. (2) An \textit{LLM Feature Factory} translates each plan into executable Python code through chain-of-thought reasoning, turning feature hypotheses into runnable programs. (3) A \textit{Self-Evolving Alignment Engine} uses Reinforcement Learning (GRPO) with dual-channel reward (metric-based utility + semantic alignment) to enhance code quality. By expressing features as programs, the learned generation patterns can transfer across domains. Evaluated on 7 public benchmarks in healthcare, finance, and social domains, Eureka consistently outperforms both traditional AutoFE and LLM-based baselines. We further demonstrate Eureka's effectiveness on cloud GPU resource demand prediction at Alibaba Cloud, where Eureka improves demand fulfillment rate by 16\% and lowers computing resource migration rates by 33\%.

\end{abstract}

\section{Introduction}
\vspace{-0.1in}

The performance of predictive models relies heavily on feature quality. Effective features are derived from variables that reflect domain logic, but creating them requires specific expertise, limiting scalability across fields. We observe that feature engineering is fundamentally a code generation problem: each feature is a program that transforms raw inputs into a predictive signal. Existing approaches treat feature design as a data problem (selecting or combining columns), but expressing features as executable code makes them composable, testable, and transferable across domains. This perspective is especially valuable in settings with sparse historical data and volatile patterns, such as cloud GPU resource forecasting~\cite{openai2024gpt4technicalreport,geminiteam2025geminifamilyhighlycapable,yang2025qwen3technicalreport}, where foundation model workloads create bursty, irregular demand that breaks conventional methods.

\begin{figure}[tbp]
    \centering
    \includegraphics[width=.9\linewidth]{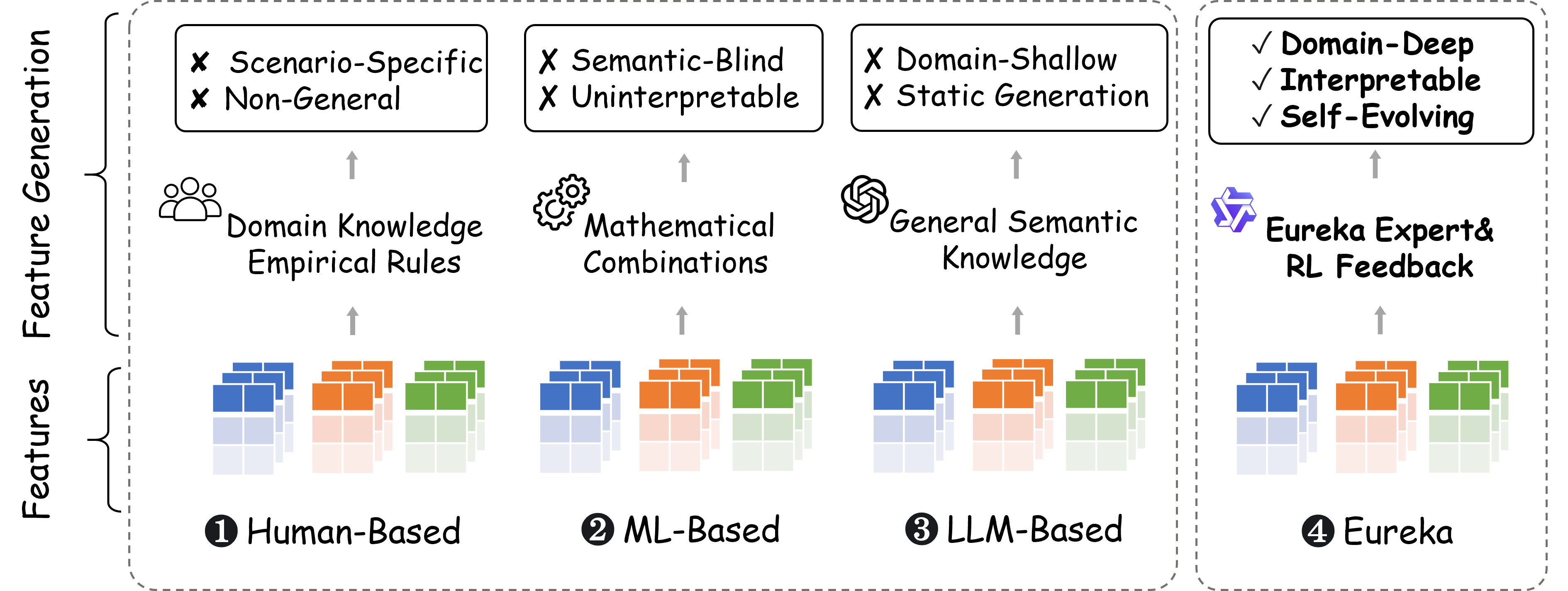}
    \caption{Comparison of different feature generation paradigms, highlighting their respective features, generation methods, and limitations/advantages.}
    \vspace{-0.15in}
    \label{fig:teaser}
\end{figure}

Across different downstreams, without domain expertise and systematic feedback for feature design, existing approaches (see Figure~\ref{fig:teaser}) remain limited:




\textbf{(1) Manual FE} relies on domain expertise, producing features effective in narrow contexts that fail to generalize across applications.

\textbf{(2) Automated FE (AutoFE)} scales feature generation but lacks semantic understanding, producing features that failed to capture bursty demand patterns~\cite{Horn2019TheAP,Kanter2015DeepFS,Zhang2022OpenFEAF}.

\textbf{(3) LLM-based FE} operates as static generators without feedback loops for post-deployment iteration, unable to adapt to irregular, bursty patterns in AI workloads~\cite{han2024large,hollmann2023large,nam2024optimized}.

To address these limitations, we introduce \textbf{Eureka}, an agentic code generation framework for feature engineering that produces executable, domain-aligned features across applications. Our system is fully integrated into production services. Our key contributions are:

\begin{itemize}
  \item \textbf{Eureka Expert (Planning Stage)} encodes domain knowledge as heuristic constraints and design templates to guide exploration of feature interactions and cross-resource dependencies. It steers toward domain-meaningful candidates, filters spurious correlations, and identifies high-order feature dependencies that reflect system dynamics.
  
  \item \textbf{LLM Feature Factory (Execution Stage)} translates design plans into concrete features via conditioned code generation. Guided by domain heuristics, it explores the feature space and outputs executable implementations that maximize information signal from limited historical data.
  
  \item \textbf{Self-Evolving Alignment Engine (Refinement Stage)} improves feature quality using dual-channel reward feedback from real-world deployment outcomes. By linking offline optimization with online performance, it adapts features to irregular workload patterns while preventing overfitting.
\end{itemize}

Together, these components form a closed-loop system where domain knowledge guides code generation, and generated code quality feeds back to improve the generation strategy. Evaluated on 7 public benchmarks spanning healthcare, finance, and social domains, Eureka consistently outperforms both traditional AutoFE and existing LLM-based methods. We further validate the framework end-to-end on cloud GPU resource demand prediction at Alibaba Cloud, where Eureka improves demand fulfillment rate by 16\% and reduces computing resource migration rates by 33\%, with 91\% of generated warnings adopted by the operations team.

\section{Related Works}
\vspace{-0.1in}

\noindent\textbf{Automated Feature Engineering.}
Feature engineering remains a critical bottleneck in machine learning deployment, requiring both statistical expertise and domain knowledge while being largely manual and experience-driven. AutoFE has emerged as a core component of AutoML to reduce iterative trial-and-error costs. FeatureTools is the most widely adopted approach, enabling automated multi-table feature fusion via Deep Feature Synthesis (DFS) \cite{Kanter2015DeepFS} for relational database. Additionally there are methods of iterative feature subsampling through bundle search to identify information-rich features \cite{Horn2019TheAP}, and feature boosting with pruning algorithms for efficient, accurate filtering \cite{Zhang2022OpenFEAF,li2023learning}.

\noindent\textbf{LLM for AutoFE.}
Recent advances in foundation models  have enabled their application to automated feature engineering \cite{hegselmann2023tabllm}. Hollmann et al. introduce a context-aware method that leverages LLMs to generate semantically meaningful features from task descriptions \cite{hollmann2023large}. Han et al. extract rule-based binary features using prior knowledge and contextual cues to support classification tasks \cite{han2024large}. Nam et al. frame feature generation as a rule synthesis problem: an LLM suggests transformation rules while a decision tree identifies the optimal ones. \cite{nam2024optimized}.

However, these methods rely on general world knowledge and lack deep domain expertises, making them less effective for complex, specialized datasets. Most importantly, they operates in open-loop manner thus cannot learn from past success or failures, often degrading to fixed combination patterns. To overcome these limitations, we propose \textbf{Eureka}, a self-evolving framework that integrates domain-guided knowledge with feedback-driven refinement.

\section{Methodology}
\vspace{-0.1in}
\begin{figure}[tbp]
    \centering
    \includegraphics[width=1\linewidth]{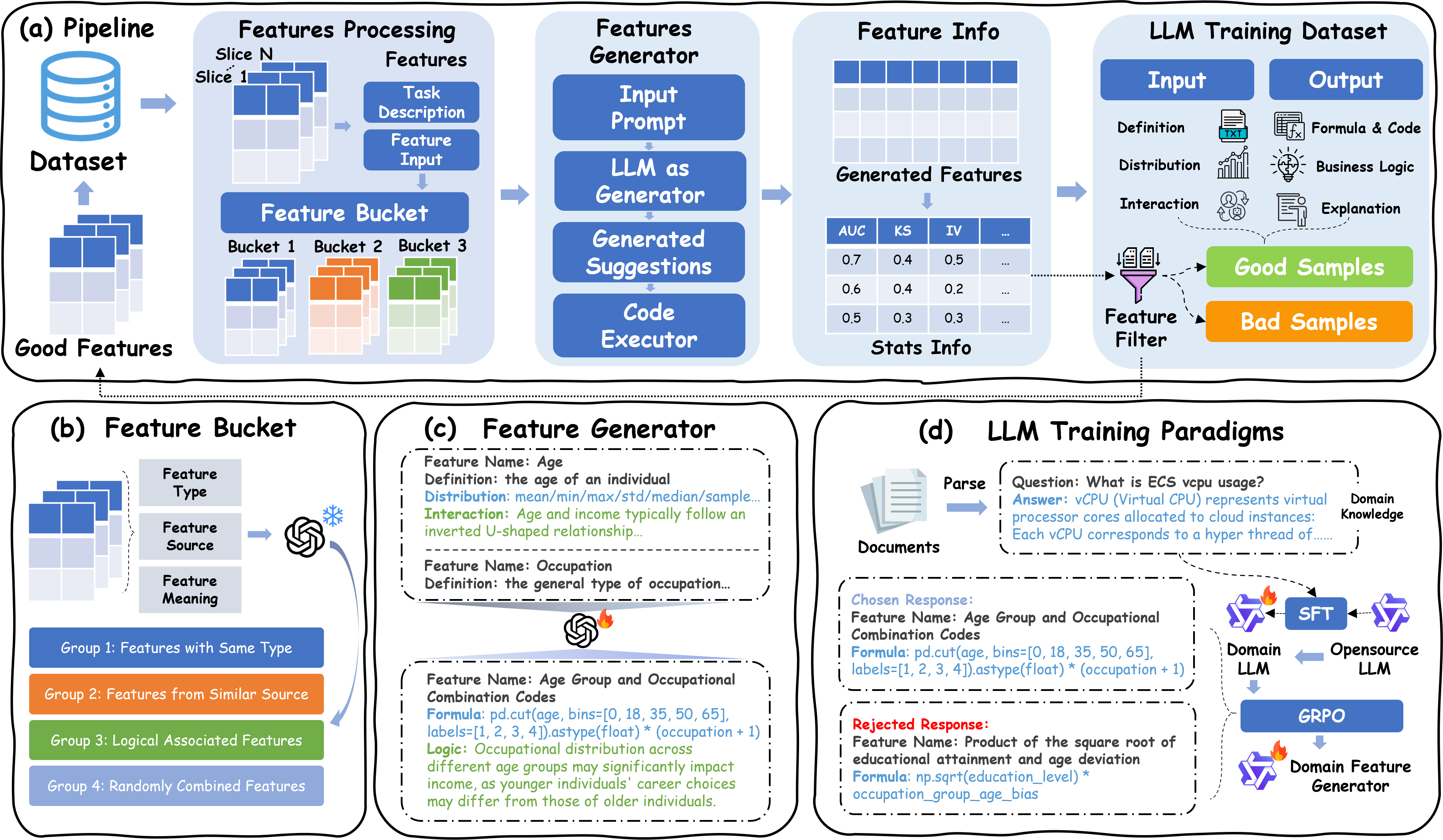}
    \caption{\textbf{Domain-specific feature generation framework.} (a) Pipeline: End-to-end workflow for generating and filtering domain-aware features. (b) Feature Bucket: Groups features by type, source, or semantic relationships. (c) Feature Generator: Uses LLMs to generate mathematical formulas and Python code for feature creation. (d) LLM Training Paradigms: Fine-tunes LLMs via SFT and GRPO for feature reasoning.}
    \vspace{-0.15in}
    \label{fig:framework}
\end{figure}
As illustrated in Figure~\ref{fig:framework}, Eureka consists of three key components: Eureka Expert, LLM Feature Factory, and Self-Evolving Alignment Engine.
\vspace{-0.1in}

\subsection{Eureka Expert}

The Eureka Expert is a domain-specialist LLM that guides principled feature exploration by encoding domain knowledge and business logic. General-purpose LLMs are insufficient in specialized business settings, so we apply supervised fine-tuning (SFT) to embed this knowledge. The open-source models has been deployed  on internal infrastructure to enable task-specific customization and ensure data security.


Our automated feature engineering is formulated as a conditional generation problem: given business context and statistical profiles of original features, the objective is to generate a structured feature design plan. This is expressed as:

\vspace{-0.08in}
\begin{equation}
    \mathit{P}^* = \arg\max_{P} \, \mathit{p}_{\theta}(\mathit{P} \mid \mathit{C}, \mathit{M})
\end{equation}
where $C$, $M$, and $P$ represent the business scenario description, feature metadata with statistical profiles, and generated feature design plan (following a predefined JSON schema), respectively. Here, $p_{\theta}(P \mid C, M)$ is the conditional probability modeled by the LLM with parameters $\theta$.

The model is fine-tuned on an expert-annotated dataset $\mathcal{D}$ to produce high-quality, domain-aligned feature design plans. Each sample includes an input $I_i$ (role definitions, task requirements, and feature statistics) and an output $O_i$ (expert-generated plan with reasoning). This process teaches structured reasoning and domain-specific construction strategies, formalized as:

\vspace{-0.08in}

\begin{equation}
    \mathcal{L}_{\mathrm{SFT}}(\theta) = - \sum_{(I_i, O_i) \in \mathcal{D}} \log p_{\theta}(O_i \mid I_i)
\end{equation}

where $\mathcal{D} = \{(I_i, O_i)\}_{i=1}^N$ is the training dataset. Minimizing $\mathcal{L}_{\mathrm{SFT}}$ encourages the model to learn expert reasoning and feature construction patterns, injecting domain-specific knowledge into its parameters. By covering diverse business scenarios and high-quality feature sets in $\mathcal{D}$, the fine-tuned model generalizes across different domains and prediction tasks. The SFT process is conducted on open-source LLMs within secure internal infrastructure, ensuring enterprise data protection. This module outputs a feature design plan with reasoning, which is passed to the LLM Feature Factory for implementation.

\begin{figure}[tbp]
    \centering
    \includegraphics[width=0.8\linewidth]{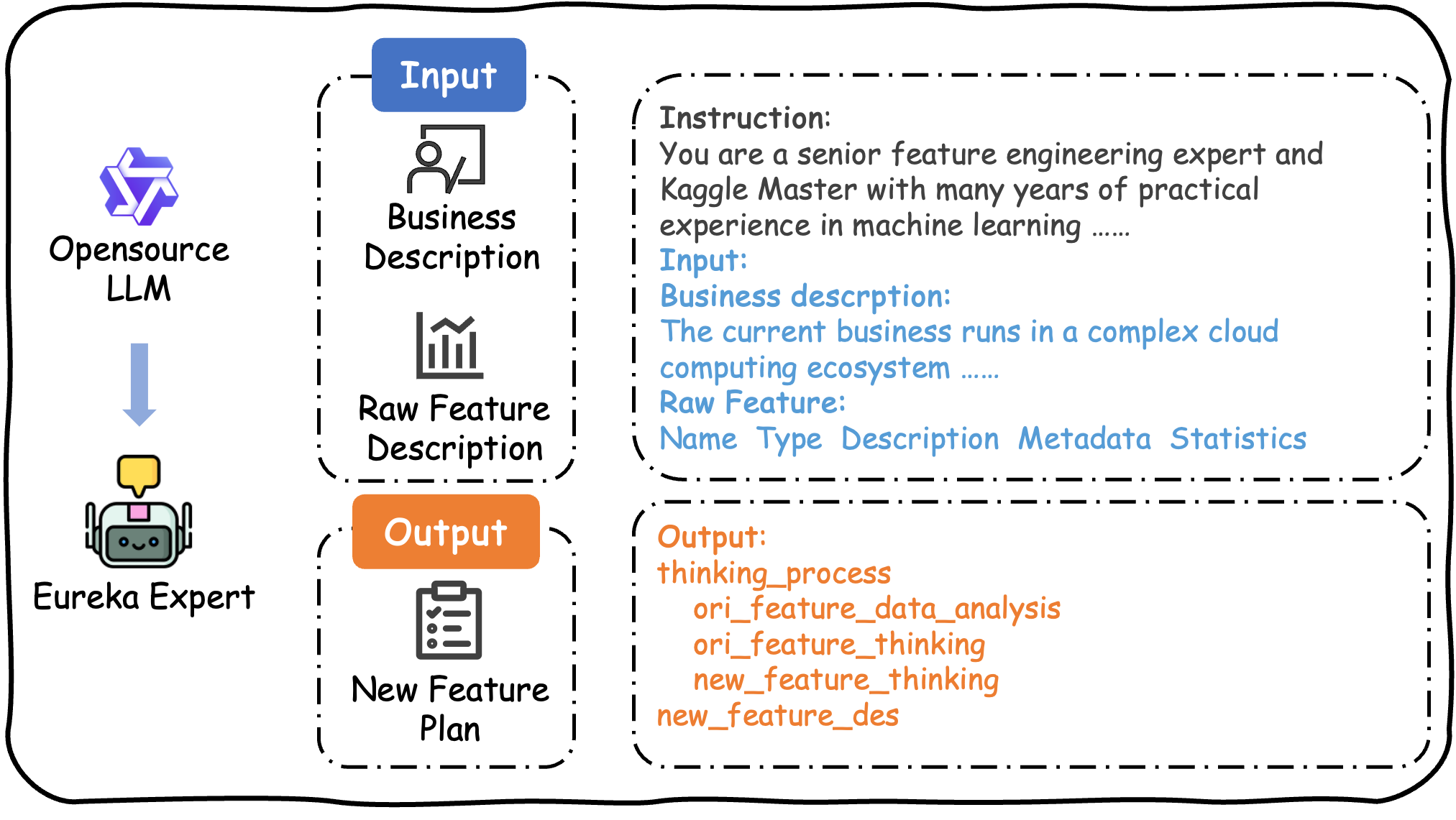}
    \caption{Eureka Expert}
    \vspace{-0.15in}
    \label{fig:placeholder}
\end{figure}

\vspace{-0.1in}
\subsection{LLM Feature Factory}
\vspace{-0.1in}
The LLM Feature Factory is the execution module that translates the feature design plans of the Eureka Expert into concrete features through code generation and validation. Feature engineering can be formulated as a bilevel optimization problem that discovers a feature $\mathcal{X}'$ through a transformation $\mathcal{F}: \mathcal{X} \rightarrow \mathcal{X}'$, where a model $f$ trained in the augmented feature space achieves better performance:
\vspace{-0.05in}
\begin{equation}
\max_{\mathcal{F}} \mathcal{\xi}(f^*; \mathcal{D}_{val} \cup \mathcal{F}(\mathcal{D}_{val})) \quad \text{s.t.} \quad f^* \in \arg\max_{f} \mathcal{\xi}(f; \mathcal{D}_{train} \cup \mathcal{F}(\mathcal{D}_{train}))
\end{equation}

Traditional gradient-based methods face challenges due to non-differentiable transformations and computational complexity. To overcome these limitations, we introduce a framework that leverages a LLM as an outer-loop optimizer, reformulating feature engineering as an iterative search process.
\begin{algorithm}[tbp]
\caption{Three-Phase LLM Feature Engineering}
\label{alg:three_phase_fe}
\begin{algorithmic}[1]
\REQUIRE Dataset $\mathbf{X}$, LLM $\mathcal{M}$, iterations $K$
\ENSURE Enhanced dataset $\mathbf{X}_{enhanced}$

\STATE \textbf{Phase 1:} $\mathcal{X} \leftarrow \text{partition}(\mathbf{X})$ into $\{\mathcal{X}_{num}, \mathcal{X}_{cat}, ...\}$
\STATE Initialize $H_0 \leftarrow \emptyset$

\FOR{$k = 1$ to $K$}
    \STATE \textbf{Phase 2:} Prepare inputs $(N_j, D_j, H_{k-1}, \text{stats}(\mathbf{X}))$
    
    \STATE \textbf{Phase 3 - Chain of Thought:}
    \STATE \quad $R_k \leftarrow \mathcal{M}.\text{generate\_rationale}(\text{inputs})$
    \STATE \quad $F_k \leftarrow \mathcal{M}.\text{formulate\_rule}(R_k)$
    \STATE \quad $C_k \leftarrow \mathcal{M}.\text{generate\_code}(F_k)$
    
    \STATE $\mathbf{x}_{new} \leftarrow \text{execute}(C_k, \mathbf{X})$
    \STATE $P_k \leftarrow \text{evaluate}(\mathbf{x}_{new})$
    
    \IF{$P_k > \text{threshold}$}
        \STATE Update: $\mathbf{X} \leftarrow \mathbf{X} \cup \{\mathbf{x}_{new}\}$, $H_k \leftarrow H_{k-1} \cup \{(N_k, D_k, P_k, S_k)\}$
    \ENDIF
\ENDFOR

\RETURN $\mathbf{X}$
\end{algorithmic}
\end{algorithm}

Our methodology, outlined in Algorithm \ref{alg:three_phase_fe}, operates in three phases. In Phase 1, the dataset is partitioned into subsets by types. Phase 2 prepares inputs including feature metadata, historical records, and statistical summaries. Phase 3 employs a Chain of Thought approach where the LLM generates a transformation rationale, formulates a feature rule, and produces executable code. New features are evaluated and good ones are retained while updating the history for subsequent iterations. This process enables exploration of the transformation space, leveraging semantic understanding to generate meaningful features.

\subsection{Self-Evolving Alignment Engine}
The Self-Evolving Alignment Engine is a reinforcement learning (RL)-based module. Its objective is to align the SFT model's outputs with human-preferred reasoning and domain-specific business context\cite{RLHF2022}, so that the model can internalize generation patterns of both high- and low-quality features and improve subsequent generalization. Given the same input context, we sample a \emph{group} of candidate feature design plans from the SFT model.
Each plan is instantiated by the LLM Feature Factory, evaluated on held-out data, and then scored by a dual-channel reward that combines
(1) metric-based utility and (2) semantic/business alignment.
We optimize the model under the GRPO framework using intra-group preference learning, which naturally matches group-wise ranking signals and
avoids training a separate reward model.

To balance feature performance and reasoning quality, the Self-Evolving Alignment Engine decomposes the reward function into a metric-based component and a semantic-based component. These are combined through weighted aggregation to produce the final optimization signal (Eq. 4).
\vspace{-0.1in}
\begin{equation}
    R_{\mathrm{total}}(P_i) = \alpha \cdot R_{\mathrm{metric}}(P_i) + \beta \cdot R_{\mathrm{semantic}}(P_i)
\end{equation}

The metric-based reward $R_{\mathrm{metric}}$ quantifies feature performance using a set of normalized statistical and modeling indicators, while the semantic-based reward $R_{\mathrm{semantic}}$ evaluates reasoning quality and domain-specific business alignment. $R_{\mathrm{metric}}$ is computed as the weighted dot product:
\begin{equation}
    R_{\mathrm{metric}}(P_i) = \mathbf{w}^T \widetilde{\mathbf{m}}_i
\end{equation}
$\mathbf{w} = [w_{\mathrm{a}}, w_{\mathrm{k}}, w_{\mathrm{i}}, w_{\mathrm{t}}, w_{\mathrm{imp}}, -w_{\mathrm{r}}]^T$ is the weight vector, and 
$\widetilde{\mathbf{m}}_i$ 
is the normalized metric vector with all metric values scaled to $[0,1]$.

In addition to metric-based performance, the semantic-based reward $R_{\mathrm{semantic}}$ measures the quality of the model's reasoning based on completeness, causal logic soundness, and alignment with business context. It is computed by employing a stronger LLM to perform group-wise 
ranking over candidates, with the resulting rank mapped to a scalar score in $[0,1]$. The reward is formally defined as:
\vspace{-0.1in}
\begin{equation}
    R_{\mathrm{semantic}}(P_i) 
    = \frac{K - \mathrm{rank}_{\mathrm{LLM}}\left(P_i \,\middle|\, \{P_j\}_{j=1}^K , \mathcal{C}\right)}{K - 1}
\end{equation}

Here, $K$ is the number of candidate features, and $\mathrm{rank}_{\mathrm{LLM}}(\cdot)$ is the rank assigned by the stronger LLM evaluator. This evaluator assesses the candidate set $\{P_j\}_{j=1}^K$ based on criteria $\mathcal{C}$: (1) reasoning completeness, (2) causal logic soundness, and (3) domain alignment. Inspired by RLAIF, we use the LLM evaluator to directly produce real-time reward signals, circumventing the need to train a separate reward model. This aligns with the GRPO framework by leveraging candidate groups and relative rankings.

To mitigate evaluator stochasticity and bias, we adopt a controlled, agreement-aware evaluation protocol. For each group, we randomly shuffle candidate order and re-evaluate for $r$ independent shuffles (we use $r{=}3$) to reduce position effects, enforce a rubric-constrained prompt that scores \emph{only} the three criteria in $\mathcal{C}$ (explicitly ignoring verbosity and stylistic flourishes), and cap/normalize the evidence shown to the evaluator with a fixed template and budget to minimize length-related bias. We aggregate the $r$ rankings via Borda count to obtain a stable group ranking, and further reduce variance using low-variance decoding. We quantify reliability using intra-evaluator agreement, computed as the average Kendall's $\tau$ across all pairs of the $r$ rankings, and check model-specific bias with a second independent evaluator, reporting inter-evaluator agreement as Spearman's $\rho$ between the two aggregated rankings. When agreement is low, we downweight the semantic reward to avoid noisy signals destabilizing RL: we scale $\beta$ according to intra-evaluator agreement $\bar{\tau}$, and disable the semantic reward when $\bar{\tau}$ is very low.

$K=5$ is set to balance preference estimation accuracy and computational cost. The weights $\alpha$ and $\beta$ controlling the trade-off between metric-based and semantic-based rewards are tuned on a held-out validation set and set to $\alpha=0.6$ and $\beta=0.4$ in our main experiments. All metric values are min–max normalized to $[0,1]$, and the semantic ranking is obtained from a GPT-4-based evaluator with a fixed evaluation prompt encoding the criteria in $\mathcal{C}$. Training employs the Adam optimizer with a learning rate of $2\times 10^{-5}$, a group batch size of $N_g=64$, and a clipping parameter of $0.2$ for stable updates.

\section{Experiment}
\vspace{-0.1in}
\subsection{Experimental Setup}

\begin{table}[tbp]
\centering
\caption{Basic information of each dataset used in our experiments. The numbers in parentheses indicate the count of categorical and numerical attributes, respectively.}
\vspace{-0.1in}
\label{tab:datasets_info}
\setlength{\tabcolsep}{12pt}
\begin{tabular}{lccc}
\hline
\textbf{Data} & \textbf{\# of samples} & \textbf{\# of features} & \textbf{Label ratio (\%)} \\
\hline
Adult         & 48842                & 14 (7/7)               & 76:24                    \\
Bank          & 45211                & 16 (8/8)               & 88:12                    \\
Blood         & 748                  & 4 (0/4)                & 76:24                    \\
Credit-g      & 1000                 & 20 (12/8)              & 70:30                    \\
Diabetes      & 768                  & 8 (0/8)                & 65:35                    \\
Heart         & 918                  & 11 (4/7)               & 45:55                    \\
Myocardial    & 1700                 & 111 (94/17)            & 22:78                    \\
EGS           & 4837                 & 45 (0/45)              & 12:88                    \\
\hline
\end{tabular}
\end{table}

\setcounter{footnote}{0}
\noindent\textbf{Datasets.} 
Evaluations are conducted on 8 classification datasets: Adult~\cite{Adult2007}, Bank~\cite{Bank2014}, Blood~\cite{Blood2009}, Credit-g~\cite{Credit2021}, Diabetes\footnote{https://kaggle.com/datasets/uciml/pima-indians-diabetes-database}, Heart\footnote{https://kaggle.com/datasets/fedesoriano/heart-failure-prediction}, Myocardial~\cite{Myocardial2020}, and a proprietary EGS (Elastic GPU Service)\footnote{Elastic GPU Service (EGS) is a cloud service from Alibaba Cloud that provides ready-to-use, scalable GPU-accelerated computing resources.} dataset, from Alibaba Cloud. The EGS dataset is derived from real GPU usage logs and used for bursty demand prediction, it is a binary classification dataset with monthly granularity. A sliding-window rate-of-change filter was then applied, followed by dual screening based on relative ranking and absolute thresholds, to ensure the validity of EGS data. We construct our supervised fine-tuning dataset using $\leq 2K$ Q\&A pairs, primarily to provide the model with domain-specific background knowledge.

\noindent\textbf{Baselines.} 
Extensive evaluations are conducted against 5 automated FE methods: Deep Feature Synthesis (DFS)~\cite{Kanter2015DeepFS}, AutoFE~\cite{Horn2019TheAP}, OpenFE~\cite{Zhang2022OpenFEAF}, CAAFE~\cite{hollmann2023large}, and FeatLLM~\cite{han2024large}, covering traditional and LLM-based approaches.

\noindent\textbf{Experimental Setup.} 
Features from all methods are combined with original features and evaluated using LightGBM~\cite{Ke2017LightGBMAH} and TabPFN~\cite{hollmann2025tabpfn}. Performance is measured by ROC-AUC, reporting the best result across both predictors. For our proposed method, each training instance was run for 100 epochs, which took 6 hours with 4 NVIDIA A100 GPUs.

\subsection{Overall Performance Comparison}
\vspace{-0.02in}

\begin{table}[tbp]
\centering
\caption{Performance Comparison Across Different Methods and Datasets}
\vspace{-0.1in}
\label{tab:performance_comparison_transposed}
\begin{tabular}{l|cccccccc}
\midrule
\textbf{Method} & \textbf{Adult} & \textbf{Bank} & \textbf{Blood} & \textbf{Credit} & \textbf{Diabetes} & \textbf{Heart} & \textbf{Myocardial} & \textbf{EGS} \\
\midrule
DFS & 0.915 & 0.897 & 0.637 & 0.707 & 0.882 & 0.921 & 0.654 & 0.680 \\
AutoFe & 0.872 & 0.865 & 0.735 & 0.676 & 0.837 & 0.903 & 0.674 & 0.666 \\
OpenFe & 0.920 & 0.923 & 0.626 & 0.701 & 0.817 & 0.897 & 0.697 & 0.658 \\
TabPFN & 0.900 & 0.904 & 0.715 & 0.787 & \underline{\textbf{0.888}} & 0.938 & 0.676 & 0.694 \\
CAAFE & 0.901 & 0.905 & 0.713 & 0.797 & 0.886 & \underline{\textbf{0.941}} & 0.686 & 0.692 \\
FeatLLM & 0.894 & 0.851 & 0.678 & 0.743 & 0.811 & 0.881 & 0.663 & 0.675 \\
\midrule
Eureka-Qwen & 0.926 & 0.932 & 0.738 & 0.794 & 0.873 & 0.926 & \underline{\textbf{0.737}} & 0.679 \\
Eureka-32B & 0.926 & 0.932 & 0.716 & 0.824 & 0.884 & 0.936 & 0.711 & 0.688 \\
Eureka-gpt4o & \underline{\textbf{0.928}} & \underline{\textbf{0.934}} & \underline{\textbf{0.745}} & \underline{\textbf{0.838}} & 0.881 & 0.938 & 0.699 & \underline{\textbf{0.699}} \\
Eureka-grok2 & \underline{\textbf{0.928}} & \underline{\textbf{0.934}} & 0.735 & 0.815 & 0.886 & \underline{\textbf{0.940}} & 0.719 & 0.686 \\
\midrule
\end{tabular}
\vspace{-0.1in}
\end{table}

Based on the results in Table~\ref{tab:performance_comparison_transposed}, several key findings emerge. The proposed Eureka framework consistently achieves strong and superior performance across multiple datasets, particularly excelling on feature-rich data such as Myocardial. This advantage stems from Eureka’s ability to semantically interpret original features and generate domain-informed feature combinations, which is especially beneficial in complex, high-dimensional settings. Eureka-Qwen, Eureka-grok2, and Eureka-gpt4o utilize corresponding proprietary APIs, while Eureka-32B is a fine-tuned open-source variant based on Qwen3-32B. Among them, Eureka-gpt4o attains the highest accuracy on four datasets (Adult, Bank, Blood, and Credit-g) and remains competitive elsewhere. Eureka-Qwen leads on Myocardial (0.737), highlighting its strength in handling feature-rich data, while Eureka-grok2 performs best on Heart and shows strong results on Adult and Bank. Overall, the consistent superiority of Eureka demonstrates the framework’s effectiveness and generalization ability in automated feature engineering.

\subsection{Ablation Study}
We conducted ablation studies to evaluate each component of Eureka on the EGS dataset by systematically removing core modules. The configurations and performance are detailed in Table~\ref{tab:ablation}.
\vspace{-0.2in}
\begin{table}[h]
\centering
\caption{Ablation study of Eureka components.}
\vspace{-0.1in}
\label{tab:ablation}
\begin{tabular}{l r r}
\toprule
\textbf{Configuration} & {\textbf{ROC-AUC (\%)}} & {\textbf{$\Delta$ vs. Eureka (\%)}} \\
\midrule
\textbf{Eureka} & 69.97 & {---} \\
\quad w/o Self Evolving Engine & 69.45 & -0.52 \\
\quad w/o Eureka Expert & 65.91 & -4.06 \\
\quad w/o LLM Factory & 63.84 & -6.13 \\
\bottomrule
\end{tabular}
\end{table}
\vspace{-0.2in}

First, the LLM factory is essential, with its removal causing the largest performance drop (6.13\%). This underscores the superiority of our automated feature discovery method compared to the traditional manual feature engineering.

Second, domain knowledge injection is critical. Replacing Eureka Expert with a generic LLM degrades performance by 4.06\% . This validates that SFT-injected domain insights are vital for generating business-relevant features. 

Finally, the self-evolving alignment provides crucial optimization. While its removal causes a 0.52\% loss, it leverages dual-channel reward feedback to guide features generation of enhanced interpretability and coherence, enabling the model to internalize successful strategies for subsequent iterations.




\subsection{Case Study}

\begin{figure}[tbp]
    \centering
    \includegraphics[width=1\linewidth]{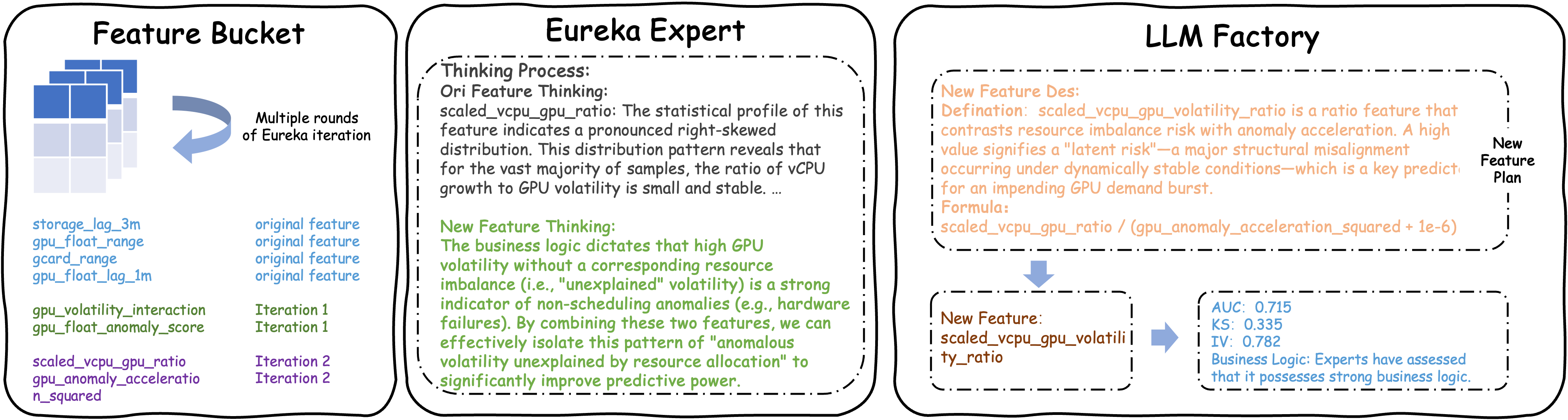}
    \caption{The process of obtaining high-quality feature}
    \label{fig:workflow}
    \vspace{-0.2in}
\end{figure}

To illustrate Eureka's practical impact, we present a production case where it successfully predicted a GPU demand burst that was undetectable using traditional monitoring systems. As illustrated in Figure~\ref{fig:workflow}, this case highlights how Eureka discovers interpretable, high-quality features by synthesizing domain knowledge rather than exhaustive feature combinations.



Initially, two weakly correlated features—gpu\_anomaly\_acceleration\_squared and scaled\_vcpu\_gpu\_ratio—were identified, which conventional methods would likely discard. Leveraging deep domain knowledge, Eureka Expert reinterpreted the elevated scaled\_vcpu\_gpu\_ratio not as a system anomaly but as a preparatory signal for large scale AI training, wherein auxiliary resources are scaled ahead of GPU intensive phases. This “preparatory imbalance,” a subtle precursor typically overlooked by statistical models, stems from customers’ proactive resource ramp‑up in anticipation of demand surges.

Guided by this business insight, the LLM Feature Factory automatically generated candidate features, and through validation by the Self Evolving Alignment Engine, the new feature scaled\_vcpu\_gpu\_volatility\_ratio was identified as optimal. This feature is engineered to precisely quantify the "preparatory imbalance" phase. It divides the high scaled\_vcpu\_gpu\_ratio (the preparation signal) by a near zero gpu\_anomaly\_acceleration\_squared (indicating a stable GPU state), amplifying the signal for pre-surge detection. This mathematical structure creates a powerful, targeted signal specifically for the "calm before the storm" scenario that precedes a major training job. In practice, the value of this new feature surges days before demand peaks while standard metrics remain flat, enabling timely forecasting from Eureka.



\section{Application in Practice}
\vspace{-0.15in}



The Eureka framework has been fully integrated in Alibaba's GPU resource management system, serving as a core module for enterprise AI computing demand forecasting. Its primary objective is to generate early warning signals of GPU demand surges through multi-source data streams.


To achieve this, Eureka creates a systematic pipeline: training data is sourced from demand records and supplemented with domain expert inputs, while the model is re-trained monthly to adapt to evolving demand patterns. Validation results from June to September 2025 demonstrate its efficacy: Eureka generated 165 warning records, with 91\% adopted by the resource operations team as actionable recommendations, confirming its business impact (detailed in Table~\ref{tab:business_impact}).





The Eureka framework has demonstrated significant business benefits in industrial applications. On the demand side, its ability to forecast demand surges for top-tier customers has increased their demand fulfillment rate by 16\%, effectively safeguarding service quality. On the supply side, Eureka has resolved the long-standing issue of high loss rates during cloud server migrations. By generating ``customer + region'' level demand forecasts, the system can reroute in-transit equipment before demand occurs, ensuring critical resources are deployed in advance while reducing unnecessary reallocations. Leveraging Eureka's migration strategy, GPU server migration rate has been reduced by 33\%, directly improving operational efficiency and reducing costs.

\begin{table}[tbp]
\centering
\caption{Quantifiable Business Impact of Eureka at Alibaba Cloud.}
\vspace{-0.1in}
\label{tab:business_impact}
\begin{tabular}{llc}
\toprule
\textbf{Impact Area} & \textbf{Key Business Metric} & \textbf{Improvement} \\
\midrule
Operational Adoption & Adoption Rate of Generated Warnings & \textbf{91\%} \\
\midrule
Demand Fulfillment & Demand Fulfillment Rate (Top-Tier Customers) & \textbf{{+}16\%} \\
\midrule
Resource Efficiency & Reduction in Server Migration Loss Rate & \textbf{{-}33\%} \\
\bottomrule
\end{tabular}
\end{table}


Eureka's advanced feature-engineering approach goes beyond traditional time-series forecasting. The framework exploits cross-feature interactions and temporal patterns for addressing complex resource demand dynamics.

Beyond immediate operational gains, Eureka has proven its ability to deliver accurate forecasts and optimize resource utilization. The framework is being extended across elastic computing to unify resource planning, enable proactive capacity management at scale, and broader applications across other domains.


\section{Conclusion}
\vspace{-0.15in}

In this paper, we address the challenge of predicting enterprise GPU demand for cloud providers in the foundation model era. We propose Eureka, an agentic feature engineering framework that incorporates domain knowledge and continuous learning. Through its three-component architecture, Eureka combines the generative power of LLMs with domain expertises. We conducted experiments on both public benchmarks and Alibaba Cloud datasets. The results demonstrate that Eureka outperforms traditional methods and existing AutoFE baselines, achieving a 4.95\% improvement in AUC on the GPU burst demand prediction.


Eureka has proven its value in business environments. Since its integration into Alibaba Cloud’s GPU resource management system, warnings generated by Eureka have achieved a 91\% adoption rate. The system has increased the demand fulfillment rate for top-tier customers by 16\%, safeguarding business stability, and reduced the GPU server migration loss rate by 33\% through forecasting, leading to improvements in resource utilization and operational efficiency.


More broadly, Eureka presents a scalable paradigm for integrating LLM-based agentic frameworks with domain knowledge in enterprise applications, paving the way toward more adaptive and knowledge-aware AI systems for supply chain and resource management. Although our evaluation centers on GPU demand forecasting, extending the framework to other resource types or domains remains a key direction. Future work also includes exploring self-supervised or adaptive reward learning to reduce reliance on human-defined heuristics.

\bibliographystyle{splncs04}
\bibliography{references}
%




\end{document}